\documentclass[letterpaper]{article} % DO NOT CHANGE THIS
\usepackage{aaai2026}  % DO NOT CHANGE THIS
\usepackage{times}  % DO NOT CHANGE THIS
\usepackage{helvet}  % DO NOT CHANGE THIS
\usepackage{courier}  % DO NOT CHANGE THIS
\usepackage[hyphens]{url}  % DO NOT CHANGE THIS
\usepackage{graphicx} % DO NOT CHANGE THIS
\usepackage{natbib}  % DO NOT CHANGE THIS AND DO NOT ADD ANY OPTIONS TO IT
\usepackage{caption} % DO NOT CHANGE THIS AND DO NOT ADD ANY OPTIONS TO IT
\usepackage{algorithm}
\usepackage{algorithmic}
\usepackage{booktabs}
\usepackage{multirow}
\usepackage{amsmath}
\usepackage{xcolor}
\usepackage{colortbl}
\usepackage{amsfonts}
\usepackage{newfloat}
\usepackage{listings}
\DeclareCaptionStyle{ruled}{labelfont=normalfont,labelsep=colon,strut=off} % DO NOT CHANGE THIS
\floatstyle{ruled}
\newfloat{listing}{tb}{lst}{}
\floatname{listing}{Listing}
\nocopyright
\title{Synth-Align: Improving Trustworthiness in Vision-Language Model with Synthetic Preference Data Alignment}
\author{
    Robert Wijaya,
    Ngoc-Bao Nguyen,
    Ngai-Man Cheung
}
\affiliations{
    Singapore University of Technology and Design\\
    8 Somapah Rd, Singapore 487372\\
    robert\_wijaya@mymail.sutd.edu.sg, thibaongoc\_nguyen@sutd.edu.sg, ngaiman\_cheung@sutd.edu.sg
}

\usepackage{bibentry}
\begin{document}

\maketitle

\begin{abstract}
Large Vision-Language Models (LVLMs) have shown promising capabilities in understanding and generating information by integrating both visual and textual data. However, current models are still prone to hallucinations, which degrade the performance and greatly harm the user experience in real-world applications. Post-training alignment, particularly preference-tuning, is intended to align model outputs and behaviors (safety, instruction-following, style), ensuring robustness and adaptability to a wide range of tasks. The use of synthetic data for alignment, particularly in multimodal settings, remains under explored. Existing approaches typically use a strong model or a ground-truth model (CLIP) to determine positive and negative image-text data points. This paper proposes SynthAlign, a pipeline to generate and collect synthetic human-preference image-text data with optimal control built specifically for post-training alignment with DPO. At the core of the framework is the utilization of reward models as a proxy of human preference. A series of evaluation and benchmarking is provided to validate the effectiveness of the proposed framework and the resulting dataset. Notably, our framework enhanced LLaVA-1.5-7B achieved substantial POPE improvements: 87.6\% accuracy and 97.8\% precision, MMHal-Bench score increased from 2.36 to 3.49, and hallucination rate decreased from 51.0\% to 25.0\% (a 50.98\% relative reduction). The model weights and dataset are available on our Hugging Face page\footnote{https://huggingface.co/datasets/pdsdpo/synthalign-v1\_0-data}.

\end{abstract}

% Uncomment the following to link to your code, datasets, an extended version or similar.
% You must keep this block between (not within) the abstract and the main body of the paper.
% \begin{links}
%     \link{Code}{https://aaai.org/example/code}
%     \link{Datasets}{https://aaai.org/example/datasets}
%     \link{Extended version}{https://aaai.org/example/extended-version}
% \end{links}

\section{Introduction}

Large vision-language models (LVLMs) \citep{Alayrac2022FlamingoAV,liu2023visualinstructiontuning,liu2024improved,Bunny,MiniGPT4,Qwen-VL,mckinzie2024mm1} have shown potential capability in understanding and generating complex information by integrating both visual and textual data, enabling advancement in downstream tasks such as caption generation and visual question answering. Despite the remarkable success, current LVLMs are still facing issue such as confidently providing mislead response in their generated output text that deviates from human preference due to the difference between the pre-training data and the distributions in real user prompts \citep{sun2023llavarlhf, yu2024rlaif, yu2024rlhf}. This discrepancy results in several problems, such as the inability to accurately capture all attributes or properties described in the text prompts, and the occasional generation of content with hallucinated details. While pretraining and supervised fine-tuning are crucial for developing foundational capabilities, post-training alignment, specifically the preference-tuning, has become an essential step before deploying LVLMs to the public to mitigate the risk of generating hallcuinate, offensive and misleading content.\\
It has long been know that higher-quality data leads to better results. Recent work shows high-quality data increase trustworthiness and reduces hallucinations in model outputs \citep{zhao2023svit, gunasekar2023textbooks}. However, the use of synthetic data for alignment, particularly in multimodal settings, remains under explored. In fact, using synthetic data gives precise control over specific image types, especially in domains where real data is scarce or difficult to obtain. In this work, we go beyond the initial foray of previous works to evaluate if high-quality, carefully curated synthetic (image-text) data could improve the SOTA vision-language models, while manage to reduce the dataset size and training compute.\\ Following the path of LLM training, reinforcement learning from human feedback (RLHF) \cite{ouyang2022training, bai2022constitutional} is a general approach to address the problems. As this technique is expensive, laborious, and also often requiring training a policy model, Direct Preference Optimization (DPO) \cite{rafailov2024direct} is proposed to improve efficiency. This method allow direct training using data obtained either from human preference or powerful LLMs. The existing approach to construct human preference typically using a strong model such as GPT-4o for preference ranking or feedback assessment \citep{2023silkie, yu2024rlaif, zhao2023hallucinations}. However, using this model for generation and ranking is neither scalable nor cost-effective for long-term. Another concurrent work \cite{ouali2024clip} constructs a dataset for DPO training using CLIP scores to generate preference pairs. While their experiments demonstrates significant improvements, we argue that the proposed dataset is relatively large for post-training purpose--which may not that efficient for DPO-based training. Additionally, there is a concern that relying solely on synthesized data, especially if it lacks quality control, can lead to a reduction in data diversity and introduce distortions \citep{Hataya_2023_ICCV, shumailov2023curse}.\\
To address the aforementioned challenges, we propose a new framework that leverages generative models and pre-trained reward models as proxies of human preferences. We employ ImageReward \cite{xu2024imagereward} and Llama-3-8B-ArmoRM scoring model \cite{wang2024interpretable} to evaluate this synthetic images and their associated responses. When noisy or uncertain data is generated, our oracle, in this case, the reward model with built-in knowledge, acts as a verifier for assessing the quality or correctness of the data. It helps decide which data points are valuable and should be retained for training, enabling us to effectively construct preferred and dispreferred response pairs for DPO training. \\
Our approach begins by generating synthetic images using Stable Diffusion\footnote{https://huggingface.co/stabilityai/stable-diffusion-3-medium} from a set of text-to-image prompts. Each generated image is evaluated by a pre-trained reward model, which rate and rank the images based on human preference alignment. The highest-ranked image is selected, and a corresponding instruction prompt is generated for further response generation using open-source LVLMs. The responses are then evaluated using a reward model that assesses helpfulness, correctness, coherence, complexity, and verbosity. The responses with the highest and lowest scores are selected as positive and negative preference data, respectively, which are then used for DPO training. By integrating these generative and reward models into our framework, we significantly reduce the human reliance on human-annotated datasets while maintaining strong alignment with human preference.\\
%Implementation Detail and results
 To validate the effectiveness of our framework and the resulting dataset, we evaluate our model across various hallucination and vision language benchmarks, demonstrating significant improvement in the trustworthiness and reasoning capabilities of the LVLMs.\\
%Contribution summary for post-training where we use a very small amount of data to get exactly the behaviors we want. Do note that small amounts of the right kinds of behavior (safety, instruction-following, style) can also make a difference, although there is a long-tail benefits from more data.
The contributions of this paper are summarized as follows:
\begin{itemize}
    \item We introduce a new pipeline for generating and collecting synthetic human-preference data with optimal control using a pre-trained reward model for a demanding post-training alignment task with DPO.
    \item We provide a small amount of synthetic human-preference image-text dataset for direct post-training alignment purposes.
    \item We conduct extensive evaluations on vision-language tasks and hallucination benchmarks to assess the performance of our model, along with some analysis of utilizing synthetic data.
\end{itemize}

\section{Related Works}

\textbf{Large Vision-Language Models.} The rapid progress in LVLMs has gained significant attention in recent years, driven by advancements in combining vision and language \citep{Alayrac2022FlamingoAV, BLIP2, instructblip, MiniGPT4, liu2023visualinstructiontuning, mckinzie2024mm1, Qwen-VL, laurenccon2024matters}. A critical factor in the success of both LLMs and LVLMs is the quality of data they are trained on. High-quality curated datasets like those used for LLMs \citep{Deita, LIMA, Alpagasus} and LVLMs \citep{Bunny, TLDR} have been crucial in enhancing language generation and image-text alignment capabilities, respectively. Recent studies have increasingly focused on refining the quality of captions for multimodal datasets to improve model performance, exploring various techniques to optimize caption quality, and emphasizing the crucial role of well-crafted captions in improving multimodal understanding \citep{chen2023sharegpt4v, chen2024allava, li2024recaption, zhao2023svit, yu2023capsfusion, lai2024veclip}. Our work builds on these advancements by introducing a new approach that leverages synthetic data and human preference modeling to further enhance the alignment and performance of LVLMs.\newline 
\textbf{Preference Alignment.} Aligning human preferences for LVLMs has become an important area of research to mitigate safety and ethical concerns in real-world applications. One of the most prominent approaches is Reinforcement Learning from Human Feedback (RLHF) \cite{ouyang2022training}, where models fine-tuned based on human preferences to better aligned with desired output. Techniques like proximal policy optimization (PPO) \cite{schulman2017proximal} and DPO     \cite{rafailov2024direct} are typically used to perform this alignment process. Recent studies have applied RLHF to LVLMs such as LLaVA-RLHF \cite{sun2023llavarlhf}. Follow-up works, such as RLHF-V \cite{yu2024rlhf} improve this by collecting human feedback in a more fine-grained form to mitigate hallucination problem in LVLMs. Subsequently, Silkie \cite{2023silkie} and RLAIF-V \cite{yu2024rlaif} explore the effectiveness of AI feedback with GPT-4 and open-source LVLMs to assess the quality of different model outputs. Recent works have utilized DPO in vision-language tasks \cite{zhou2024aligning} and incorporated CLIP for scoring samples \cite{ouali2024clip}. While their experiments showed significant improvements, the proposed dataset's size relatively large. Our resulting dataset comprises only 9K image-text pairs, leveraging a human preference reward model as a proxy for human feedback to ensure higher quality, responsibile, and safety alignment in downstream tasks.

\section{SynthAlign: Synthetic Preference Data for Post-Training Alignment with DPO}
\begin{figure*}
    \centering
    \includegraphics[width=\linewidth]{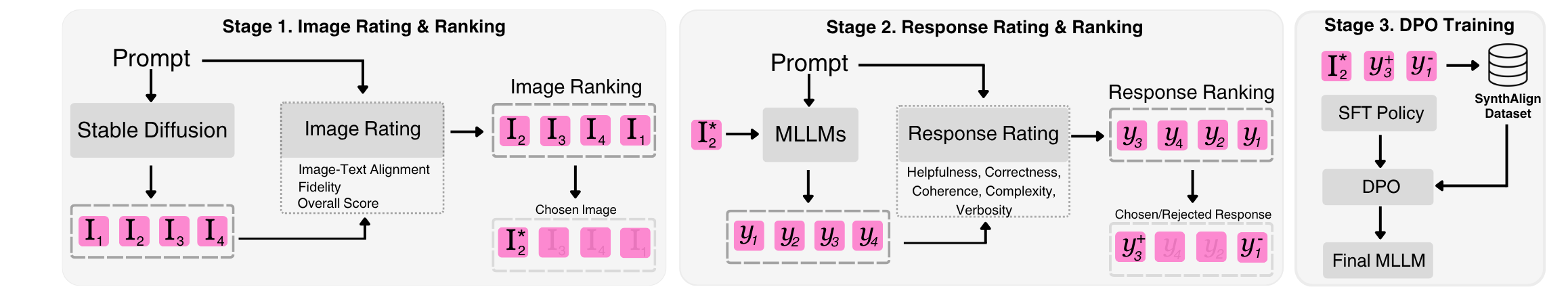}
    \caption{Overview of the framework: Starting with an initial text-to-image prompt, the Stable Diffusion model generates synthetic images. These images are then filtered using a reward model to exclude low-quality samples and retain only those with the highest scores. The selected images, along with their corresponding instruction prompts, serve as input for open-source LVLMs to generate responses. These responses are evaluated based on various criteria, and only the highest-scoring ones are selected to identify the most suitable positive and negative pairs for DPO-based training.}
    \label{fig:framework}
\end{figure*}

 The framework follows a typical DPO-based training pipeline and consists of three key stages: generation, annotation, and optimization. We introduce modifications that leverage generative and reward models; specifically, our approach consists of: (1) image generation and ranking, and (2) response generation and ranking, (3) DPO training. Initiating with a GPT-4-derived prompt, we employ Stable Diffusion to synthesize four different images, each with a different guidance scale. A pre-trained reward model evaluates these images, and we select the highest-scoring one. Subsequently, both the selected image and the initial text-to-image prompt serve as the basis for crafting an instruction prompt. These responses are assessed by a reward model based on helpfulness, correctness, coherence, complexity, and verbosity. The top-scoring response is chosen as positive preference data, and the lowest as negative preference data. Finally, the model undergoes DPO training using Equation \ref{Eq: DPO_loss_funtion}, as illustrated in Figure \ref{fig:framework}.
\subsection{Background}
\textbf{Direct Preference Optimization.} Direct Preference Optimization (DPO) is a framework designed for preference learning, where the goal is to optimize a model's behavior based on human preferences. The theoretical foundation of DPO is the Bradley-Terry model \cite{bradley1952rank} that used to quantify how well the learned reward function aligns with the preference data. \\
We consider a dataset of preference pairs 
$\mathcal{D}_P = \{x^{(i)}, y_{w}^{(i)}, y_{l}^{(i)}\}_{i=1}^{N}$ where each $x^{(i)}$ represents an input prompt, in this case, a combination of text and images modality, and $y_w^{(i)}$ and $y_l^{(i)}$ denote the preferred and dis-preferred responses to that corresponding to the input $x$, respectively. The probability that a response $y_w$ is preferred over $y_l$ is given by:
\begin{equation}
    P(y_w > y_l) = \sigma(r_\phi(x, y_w) - r_\phi(x, y_l))
\end{equation}
This equation reflect the Bradley-Terry model's assumption that the probability of preferring one response over another depends on the difference in their assigned rewards. The goal is to train a reward function $r_\phi(x,y)$ that assigns higher scores to preferred responses over dis-preferred ones. This can be achieved by minimizing the following loss function.
\begin{equation}
    \mathcal{L}_{RM}(\phi) = - \mathbb{E}_{(x, y_w, y_l) \sim \mathcal{D}_P} [ \log \sigma(r_\phi(x, y_w) - r_\phi(x, y_l))]
\end{equation} where $\sigma$ denotes the sigmoid function. The next step is we duplicate the policy model $\pi_{\theta}$ to create a reference model $\pi_{ref}$ which remains frozen during training. The reference model serves as a baseline, helping to regularize the updated policy and prevent it from deviating too far from the initial model. The overall objective is formulated as follows.
\begin{equation}\label{eq:obj}
    \max_{\pi_\theta} \mathbb{E}_{x \sim \mathcal{D}, y \sim \pi_\theta(y|x)} [r_\phi(x, y)] 
    - \beta \mathbb{D}_{KL}[\pi_\theta(y|x) || \pi_{ref}(y|x)]
\end{equation} where $\beta$ is the KL penalty coefficient, and $\mathbb{D}_{KL}$ denotes the Kullback-Leibler divergence between the updated policy $\pi_{\theta}$ and reference policy $\pi_{ref}$. By leveraging the structure of the objective in Equation \ref{eq:obj}, we can reformulate the optimization as a classification loss over the preferred data, which leads to the following loss function for DPO.  
% \hspace{5mm}$\mathcal{L}_{DPO}(\pi_{\theta};\pi_{ref};\mathcal{D}_{P}) 
%     &= -\mathbb{E}_{(x, y_w, y_l) \sim \mathcal{D}_P} $
% \begin{align}
%     &\left[ \log \sigma \left( \beta \log \dfrac{\pi_\theta(y_{w}|x)}{\pi_{ref}(y_w|x)} - \beta \log \dfrac{\pi_\theta(y_{l}|x)}{\pi_{ref}(y_l|x)} \right) \right]
% \end{align}
\begin{align}
    \label{Eq: DPO_loss_funtion}
    &\quad \mathcal{L}_{DPO}(\pi_{\theta};\pi_{ref};\mathcal{D}_{P}) 
    = -\mathbb{E}_{(x, y_w, y_l) \sim \mathcal{D}_P} \nonumber \\
    &\left[ \log \sigma \left( \beta \log \dfrac{\pi_\theta(y_{w}|x)}{\pi_{ref}(y_w|x)} 
    - \beta \log \dfrac{\pi_\theta(y_{l}|x)}{\pi_{ref}(y_l|x)} \right) \right]
\end{align} This loss function transforms the problems into a binary classification task, where the model learns to assign higher probabilities to preferred responses compared to dispreferred ones.
\subsection{Image Generation and Ranking}
\begin{figure}
    \centering
    \includegraphics[width=1\linewidth]{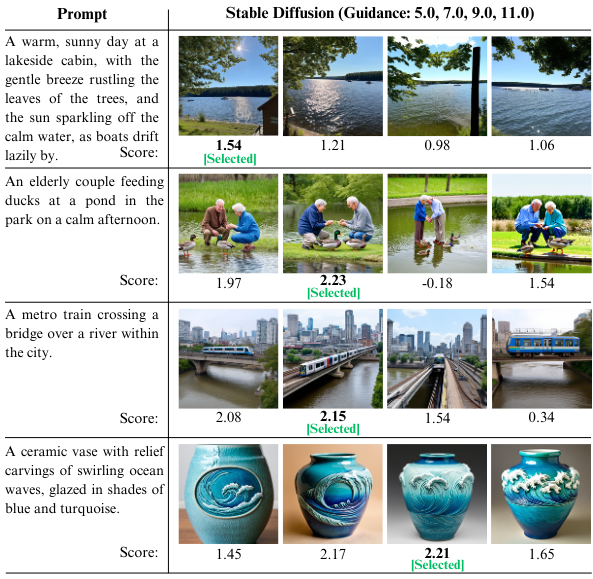}
    \caption{The figure illustrates image generation results using Stable Diffusion across four different guidance scales (5.0, 7.0, 9.0, 11.0), with the highest-scored image selected for each prompt based on a preference model evaluation.}
    \label{fig:stable-diffusion}
\end{figure}

We utilize GPT-4 to generate text-to-image prompts spanning 10 different topics, including art, school, transport, weather, and daily activities ensuring diversity in content. Let $\mathcal{P} = \{p_1, p_2, ..., p_n\}$ denote the set of prompts, where $p_i$ represents the $i$-th prompt in the set. For each prompt $p_i$, we use Stable Diffusion to generate synthetic images. By adjusting the guidance scale, we generate a set of four images $\mathcal{I}_{p_i} = \{I_1, I_2, I_3, I_4\}$ for each prompt $p_i$. \\
Let $I_{p_i}^g$ represent the generated image for prompt $p_i$ with guidance scale $g$. Formally we can express the image generation as follows:
\begin{equation}
    I_{p_i}^g = SD(p_i, g) 
\end{equation}
where $g$ varies for each image to achieve different styles and levels of prompt alignment. This approach helps generate diverse outputs for each prompt while maintain consistency in the generation process. We conduct an analysis specifically on guidance scale (see Section \ref{analysis}) to study scale at which the model produces preferable images. To reduce noise and enhance robustness, the generated synthetic images are filtered to remove low-quality samples. An image filtering strategy, serving as a proxy for human preferences, is employed to exclude low-quality images and retain those that are more likely be preferred by humans. Specifically, each image in $\mathcal{I}_{p_i}$ is evaluated using a pre-trained reward model \cite{xu2024imagereward}, denote as $R_{img}(I)$ trained on human feedback to assign quality scores to the synthetic images. The reward model evaluates the images according to criteria such as image-text alignment and fidelity. The score for each image is denoted as follows:
\begin{equation}
    s_{I_j} = R_{img}(I_j)
\end{equation}
where $I_j \in \mathcal{I}_{p_i}$. Images with lower scores are discarded, while image with the highest score is selected as the representative image:
\begin{equation}
    I_{p_i}^* = \arg\max_{I_j \in 
 \mathcal{I}_{p_i}} s_{I_j}
\end{equation}
Unlike CLIP which measures image-text similarity through contrastive alignment, the reward model provides a more nuanced evaluation by considering multiple criteria that reflects human preferences. By integrating reward model as our filtering mechanism, we achieve a higher degree of quality control in image selection. This approach surpasses CLIP's capabilities in tasks where image quality, fidelity, and nuanced alignment with text prompts are critical for constructing synthetic dataset. \\
The selection process for generated images involves assigning each image a score by the reward model, with the highest-scoring image being selected. For example, in the second row of Figure \ref{fig:stable-diffusion} (elderly couple feeding ducks), the image generated with a guidance scale of 7.0 received the highest score (2.23), likely due to its alignment with the prompt, where the couple is seated close to the ducks. In contrast, the third image, generated with a guidance scale of 9.0, received the lowest score (-0.18) because it depicts an unrealistic scenario where the elderly couple appears to be standing in the water and does not align well with the prompt description. Similarly, in the third row (metro train crossing a bridge), the first and second images have relatively small score differences (2.08 and 2.15), as both are visually appealing and closely match the prompt's description of a cityscape with a train, though the second image slightly better captures the intended details of the scene. This evaluation demonstrates how the reward model helps select the most relevant image from multiple outputs, optimizing the use of Stable Diffusion for text-to-image tasks.
\begin{figure*}
    \centering
    \includegraphics[width=\linewidth]{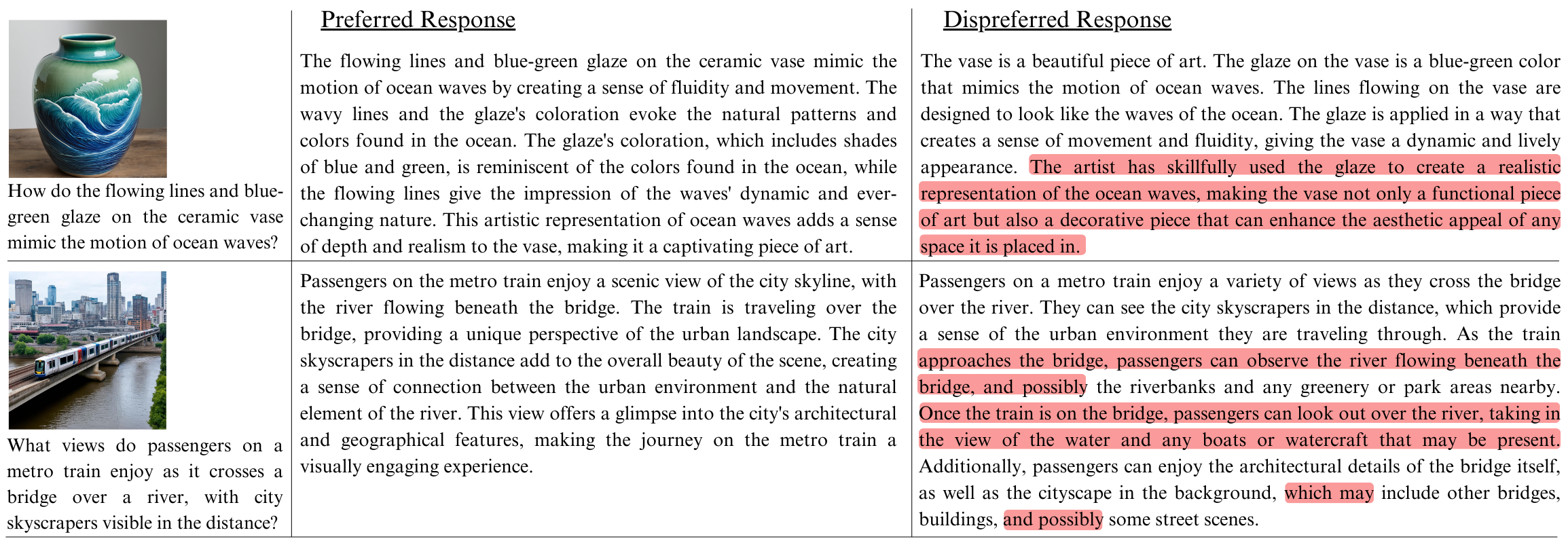}
    \caption{\textbf{Sample of preferred and dispreferred responses.} The dispreferred ones include speculative, redundant, or unclear information. This highlights the quality distinctions in vision-language task performance in the context of preference learning.}
    \label{fig:figure-02}
\end{figure*}
\subsection{Response Generation and Ranking}
Given the selected image $I_{p_i}^*$, we generate a corresponding instruction prompt $q_{p_i^*}$ utilizing GPT-4o. This is accomplished by extracting the relevant information from the original text-to-image prompt and crafting an instruction prompt designed to capture the specific details of the image. This instruction prompt along with the image, is then fed as input to open-source LVLMs, which generate responses $\mathcal{Y}_{p_i}=\{y_1, y_2, y_3, y_4\}$ from different number of LVLMs, including LLaVA-1.6-Mistral-8x7B, LLaVA-1.6-Vicuna-13B, and LLaVA-1.6-Llama-8B, InternVL2.5-8B, and Mini-InternVL-4B. Let the response generation process be represented as follows:
\begin{equation}
    \mathcal{Y}_{p_i} = LVLMs(I_{p_i}^*, q_{I_{p_i^*}})
\end{equation}
where $\mathcal{Y}_{p_i}$ denotes the set of four generated responses for prompt $p_i$ and the chosen image $I_{p_i}^*$. Each response $y_j \in \mathcal{Y}_{p_i}$
 is evaluated by a reward model $R_{resp}(y_j)$, in this case we use Llama-3-8B-ArmoRM \cite{wang2024interpretable} to produce score of each response based on helpfulness, correctness, coherence, complexity, and verbosity.
 \begin{equation}
     s_{y_j} = R_{resp}(y_j)
 \end{equation}
 The response with the highest score is selected as the positive preference data $y^+$, and the response with the lowest score is selected as the negative preference data $y^-$, denote as follows.
 \begin{equation}
     y^{+} = \arg\max_{y_j \in \mathcal{Y}_{p_i}} s_{y_j}, \quad y^{-} = \arg\min_{y_j \in \mathcal{Y}_{p_i}} s_{y_j}  
 \end{equation}
 This expression represents selecting the response $y^+$ with the highest score and the response $y^-$ with the lowest score.
 The dispreferred responses highlighted in Figure \ref{fig:figure-02} demonstrate issues such as redundancy, speculative language, and lack of precision. For instance, in the dispreferred response about the vase (top row), unnecessary repetition can reduces clarity, while speculative details in the metro scene (bottom row) reduce reliability and coherence. Conversely, the preferred responses are more concise, focus on key visual elements, and articulate relevant details with precision. The reward model, Llama-3-8B-ArmoRM \cite{wang2024interpretable}, evaluates these characteristics, ensuring the selection of responses that are helpful, accurate, and engaging, improving the overall quality and relevance of the content generated by LVLMs.
\subsection{Discussion}
Our framework leveraging synthetic data generated through generative models and evaluated using pre-trained reward models as proxies for human preferences. This allows us to effectively filter and select high-quality data for training, addressing issues related to data quality and alignment. Notably, our analysis reveals that our reward models shows significant overlap with methods like CLIP and BLIP in selecting optimal images, indicating aligned preferences in image-text alignment and visual quality (See Table \ref{common-selection}). However, when we perform LLM-as-a-Judge with stronger model (GPT-4o) in order to select the best image given two images and a caption, the Reward Model (RM) performs better than CLIP in image-text alignment tasks. Out of 100 total comparisons, RM won 53 times, CLIP won 37 times, and there were 10 ties (See Table \ref{table:outcomes}).
% Additionally, we observe that our approach demonstrate a strong balance between coherence and diversity in generated responses compared to LLaVA-RLHF and RLAIF-V, consistently producing more detailed prompts and responses (see Figure \ref{fig:stable-diffusion}). However, prior works could not achieve such improvements, often facing challenges related to data diversity and potential distortions due to lack of quality control in synthetic data. More detailed results are provided in Section \ref{analysis}.
\section{Experiments}
In this section, we conduct evaluations on visual language tasks and hallucination benchmarks, as detailed in Section \ref{mainresults}. Also, we analyze the effectiveness of synthetic data and the proposed DPO framework, as discussed in Section \ref{analysis}.
\subsection{Experimental Settings}
\textbf{Implementation Details.} In this work, we utilize the LLaVA-1.5 and LLaVA-1.6 architecture, which utilize CLIP vision encoder \cite{radford2021learningtransferablevisualmodels}. We keep the vision encoder frozen during training. For the LLM, we use the pre-trained Vicuna 7B \cite{Qwen-VL} a LoRA \cite{hu2021lora}. The models are subsequently fine-tuned for 2 epochs using DPO-based optimization on the curated dataset. We set gradient accumulation steps are set to 4, and learning rate of 1e-5. Model training is performed using DeepSpeed Zero-2 on 2 $\times$ 80GB A100 GPUs.\\
\textbf{Evaluation Benchmarks.} To assess the performance of our LVLMs, we evaluate them on various vision-language tasks and hallucination benchmarks. For vision language tasks, we adopt VizWiz \cite{gurari2018vizwiz}, MMMU \cite{yue2024mmmu}, SEED \cite{li2023seed}, SQA \cite{lu2022learn}, MME \cite{fu2023mme}, MM-Vet \cite{yu2023mm}. Each of these benchmarks contains task that evaluate reasoning, cognition, and perception capabilities of LVLMs. These evaluation results for general vision language tasks are obtained using standardized framework, LMMs Eval \cite{zhang2024lmmsevalrealitycheckevaluation}.\\
We further evaluate our approach using comprehensive hallucination benchmarks, including Object HalBench \cite{rohrbach2018object}, AMBER \cite{wang2023llm}, and POPE \cite{li2023evaluating}. Briefly, Object HalBench is used to assess common object hallucinations in image descriptions; MMHal-Bench asks GPT-4V to compare model outputs with human responses to evaluate response-level hallucination rates and informativeness; and AMBER is an LLM-free, multidimensional benchmark for LVLM hallucination evaluation.\\
\textbf{Baselines.} We first compare our proposed approach on visual-language tasks proposed approach on visual language tasks with the general-purpose LVLMs, including BLIP-2 \cite{BLIP2}, Instruct-BLIP \cite{instructblip}, Shikra \cite{chen2023shikra}, MiniGPT-4 \cite{MiniGPT4}, Qwen-VL \cite{Qwen-VL}, and LLaVA-1.5 \cite{liu2024improved}. Additionally, we assess our model on hallucination benchmarks against other preference-learning LVLMs, such as LLaVA-RLHF \cite{sun2023llavarlhf}, Silkie \cite{2023silkie}, RLHF-V \cite{yu2024rlhf}, and CLIP-DPO \cite{ouali2024clip}.
\subsection{Main Results}\label{mainresults}
\textbf{Evaluation on Visual-Language Tasks and Hallucination Benchmarks.} The evaluation results on visual language tasks are reported in Table \ref{results-vqa}. We observed that our framework sets a new benchmark in vision-language tasks among other open-source models. Notably, our models demonstrate competitive score in several benchmarks, including SQA, MM-Vet, and MMBench. Additionally, the evaluation results on hallucination benchmarks in Table \ref{tab:results-hallucination} showing improvement in hallucination control and accuracy, particularly reflected in the MMHal-Bench and POPE score.
\begin{table*}[t]
% Save the current value of \tabcolsep
%\newlength{\oldtabcolsep}
\newlength{\oldtabcolsep}
\setlength{\oldtabcolsep}{\tabcolsep}

% Set custom \tabcolsep just for this table
\setlength{\tabcolsep}{10pt} 
\renewcommand{\arraystretch}{0.8}% Adjust the value as needed
\begin{center}
\centering
\scriptsize
\setlength{\tabcolsep}{10pt}
\resizebox{\textwidth}{!}{ % Adjusted resize to only scale width
\begin{tabular}{@{}l l l c c c c c c@{}}
\midrule
\multirow{2}{*}{Model} & \multirow{2}{*}{Dataset Size} & 
\multirow{2}{*}{Feedback} & \multicolumn{2}{c}{\textsc{MMHal-Bench}} & \multicolumn{2}{c}{\textsc{AMBER}} & \multicolumn{2}{c}{\textsc{POPE}}\\ 
% \cline{4-5} \cline{6-7} \cline{8-8}% Adjusted clines for removed column
&  &  & Score $\uparrow$ & Hall. $\downarrow$ & CHAIR $\downarrow$ & Cover $\uparrow$ & Acc. $\uparrow$ & Pre. $\uparrow$\\
\midrule
Qwen-VL \cite{Qwen-VL} &  &  & 2.76 & 38.5 & 6.6 & 53.2  & - & -\\
GPT4-V &  & Unspecified & 3.49 & 28.1 & 4.6 & \textbf{67.1} & - & -\\
LLaVA-RLHF \cite{sun2023llavarlhf} & 122k & Human & 2.02 & 62.5 & 9.7 & 53.2 & 81.5\% & 87.2\%\\
% LLaVA-NeXT \cite{li2024llava} & Qwen 7B & - & 3.31 & 34.4 & 81.4 & 85.4  \\
\midrule
LLaVA-1.5-7B \cite{liu2024improved} &  &  & 2.36 & 51.0 & 7.7 & 51.6 & 86.1\% & 89.1\%\\
+ RLHF-V \cite{yu2024rlhf} & 1.4k & Human & 2.45 & 51.0 & 6.3 & 46.1 & 86.2\% & -\\
+ Silkie \cite{2023silkie} & 40k & GPT-4V & 3.19 & 32.3 & 5.4 & 55.8 & 83.7\% & - \\
+ POVID \cite{zhou2024aligning} & 17k & GPT-4V & 2.08 & 56.7 & 7.4 & 51.3 & 86.9\% & 89.0\%\\
+ RLAIF-V \cite{yu2024rlaif} & 16k & MLLM & 3.06 & 29.2 & \textbf{3.0} & 50.4 & 81.6\% & 94.6\%\\
+ CLIP-DPO \cite{ouali2024clip} & 750k & CLIP & - & - & 3.7 & 47.8 & 85.8\% & - \\
\rowcolor{gray!10} + SynthAlign (ours) & 9k & RM & \textbf{3.49} & \textbf{25.0} & 7.5 & 49.7 & \textbf{87.6\%} & \textbf{97.8\%}\\
% \midrule
% LLaVA-1.6-7B \cite{liu2024llavanext} &  &  &  &  &  &  & 86.5\% & \\
% \rowcolor{gray!10} + PDS-DPO (ours)& 12k & RM &  &  &  &  & \textbf{87.6\%} & \textbf{98.2\%}\\

% \midrule
% LLaVA-1.5 \cite{liu2024improved} & Vicuna 13B & - & - & - & 3.44  & - & - & - & 86.2\\
% \rowcolor{gray!10} + PDS-DPO & Vicuna 13B & RM & - & - & - & - & - & - & -\\
\midrule
\end{tabular}
}
\end{center}
\vspace{-10pt}
\caption{Hallucination evaluation results on several benchmarks. The inclusion of our approach with LLaVA-1.5-7B models shows performance improvements.}
% Reset \tabcolsep to its original value
\label{tab:results-hallucination}
\setlength{\tabcolsep}{\oldtabcolsep}
\end{table*}

\begin{figure}[h]
    \centering
    \includegraphics[width=1\linewidth]{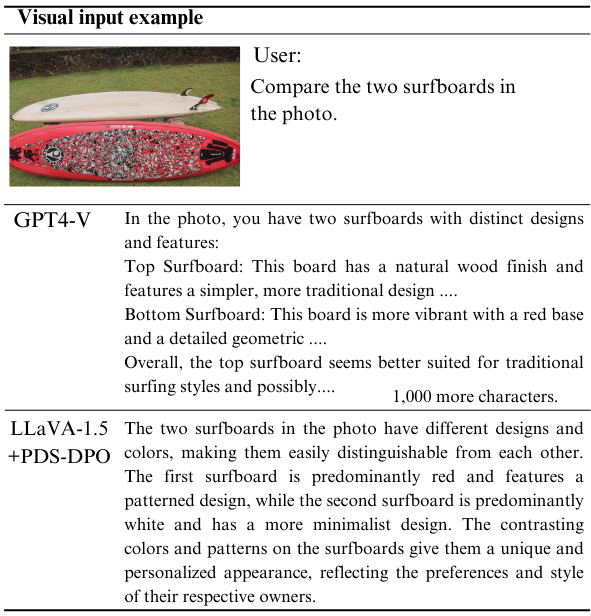}
    \caption{Qualitative demonstrations comparing our method with GPT4-V.}
    \label{fig:qualitative}
\end{figure}

\begin{table*}
% Save the current value of \tabcolsep
% \newlength{\oldtabcolsep}
\setlength{\oldtabcolsep}{\tabcolsep}
% Set custom \tabcolsep just for this table
\setlength{\tabcolsep}{10pt} % Adjust the value as needed
\renewcommand{\arraystretch}{1.0}
\begin{center}
\resizebox{\textwidth}{!}{
\begin{tabular}{l l | l l l l l l}
\midrule
Model  & LLM & ScienceQA & MME & VizWiz & MMMU & SEED & MM-Vet \\
\midrule
BLIP-2 \cite{BLIP2}& Vicuna 13B & 53.8 & 290.0 & 25.3 & 34.4 & 49.7 & 22.4\\
Intruct-BLIP \cite{instructblip}& Vicuna 13B & 63.1 & 291.8 & 34.5 & 32.9 & 57.8 & 26.2\\
Shikra \cite{chen2023shikra}& Vicuna 13B & 45.8 & - & - & - & - & -\\
MiniGPT-4 \cite{MiniGPT4}& Vicuna 7B & 42.8 & 144.3 & - & 26.8 & 47.4 & -\\
Idefics2 \cite{laurenccon2024matters}& LLaMA 7B & - & - & 35.5 & - & 44.5 & -\\
Qwen-VL \cite{Qwen-VL}& Qwen 7B & 67.1 & - & 35.2 & - & 62.3 & -\\
Qwen-VL-Chat \cite{Qwen-VL}& Vicuna 7B & 68.2 & 360.7 & 38.9 & 35.9 & 65.4 & -\\
Intern-VL-Chat \cite{chen2024internvl}& Vicuna 7B & - & 341.1 & 52.5 & \textbf{39.1} & - & -\\
\midrule
LLaVA-1.5 \cite{liu2024improved} & Vicuna 7B & 66.8 & 355.7 & 50.0 & 35.9 & 66.1 & 31.1\\
\rowcolor{gray!10}
+ SynthAlign-9K (ours) & Vicuna 7B & 69.0\textsubscript{\textcolor{teal}{\scriptsize +3.3\%}} & 367.5\textsubscript{\textcolor{teal}{\scriptsize +3.3\%}}& \textbf{55.3}\textsubscript{\textcolor{teal}{\scriptsize +10.6\%}} & 36.1\textsubscript{\textcolor{teal}{\scriptsize +0.6\%}}& \textbf{66.3}\textsubscript{\textcolor{teal}{\scriptsize +0.3\%}} & \textbf{34.1}\textsubscript{\textcolor{teal}{\scriptsize +9.6\%}}\\
\rowcolor{gray!10}
+ SynthAlign-12K (ours) & Vicuna 7B & \textbf{69.4}\textsubscript{\textcolor{teal}{\scriptsize +3.9\%}} & \textbf{378.2}\textsubscript{\textcolor{teal}{\scriptsize +6.3\%}} & 55.2\textsubscript{\textcolor{teal}{\scriptsize +10.4\%}} & \textbf{36.2}\textsubscript{\textcolor{teal}{\scriptsize +0.8\%}} & \textbf{66.3}\textsubscript{\textcolor{teal}{\scriptsize +0.3\%}} 
& 33.4\textsubscript{\textcolor{teal}{\scriptsize +7.4\%}}\\

\midrule
\end{tabular}
}
\end{center}
\vspace{-10pt}
\caption{Performance comparison of LVLMs across general benchmarks. It highlights the improvements achieved by incorporating our method with LLaVA-1.5-7B models.}
% Reset \tabcolsep to its original value
\label{results-vqa}
\setlength{\tabcolsep}{\oldtabcolsep}
\end{table*}
\hspace{-0.35cm}\textbf{Qualitative Results.} For qualitative results, Figure \ref{fig:qualitative} provides visual examples demonstrating the differences in GPT4-V, and our model outputs. The figure illustrate our model's capability to keep concise and relevance in answer while remains details. Additionally, the comparison highlights how our model able to provide more contextually appropriate response compared to counterparts that tend to generate lengthy answer.
\subsection{Analysis}\label{analysis}
\begin{figure}[h]
    \centering
    \includegraphics[width=0.90\linewidth]{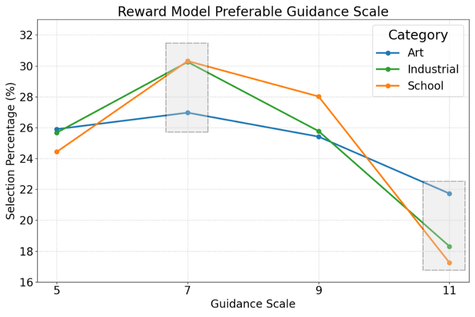}
    \caption{Selection percentage of preferred guidance scales by the reward model.}
    \label{fig:preferable-guidance}
\end{figure}

\textbf{Preferable Guidance Scale.} We observed that adjusting the guidance scale significantly impacts the quality of synthetic data, as determined by the preferences picked by the reward model. As shown in Figure \ref{fig:preferable-guidance}, we sampled three image categories: Art, Industrial, and School related images, and found that a guidance scale of 7.0 yielded the highest selection percentages, peaking at around 30\%. However, when the guidance scale increased to 11.0, there was a sharp decline in preference, with all categories dropping below 20\%, indicating that higher guidance scales may compromise visual coherence. In contrast, a guidance scale of 5.0 produced moderate selection percentages. This highlighting the necessity of carefully choosing the image candidates to improve the overall effectiveness of models trained on this synthetic data. 
\begin{table}[t]
\setlength{\oldtabcolsep}{\tabcolsep}
\setlength{\tabcolsep}{5pt} % Adjust the value as needed to reduce padding
\begin{center}
\resizebox{0.45\textwidth}{!}{ % Adjusted width to half page
\begin{tabular}{l c c c c c} % Fixed column count
\toprule
\multirow{2}{*}{Model} & \multicolumn{2}{c}{\textsc{Obj.Hal Bench}} & & \multicolumn{2}{c}{\textsc{POPE}} \\ % Added missing braces
\cmidrule{2-3} \cmidrule{5-6} % Fixed cline syntax and column ranges
& $\text{CHAIR}_{\textit{S}}$↓ & $\text{CHAIR}_{\textit{i}}$↓ & & Acc. $\uparrow$ & Pre. $\uparrow$\\
\midrule
Qwen-VL-Chat-7B & 48.2 & 9.1 & & \textbf{87.07\%} & - \\
LLaVA-1.5-7B & 66.8 & 12.7 & & 85.90\% & 89.10\% \\
+ HA-DPO & 54.0 & 14.2 & & 84.90\% & \textbf{90.42\%} \\
+ VLFeedback & 56.3 & 11.4 & & 83.72\% & - \\
+ POVID & 50.7 & 15.3 & & 84.77\% & 89.01\% \\
+ RLHF & 54.0 & 9.3 & & 81.50\% & 87.20\% \\
+ RLHF-V & \textbf{44.6} & \textbf{7.9} & & 86.20\% & - \\
\midrule
\rowcolor{gray!10} +SynthAlign-2K & 47.3 & 35.9 & & 82.10\% & 96.75\% \\
\rowcolor{gray!10} +SynthAlign-5K & 43.7 & 28.0 & & 84.77\% & \textbf{97.91\%} \\
\rowcolor{gray!10} +SynthAlign-7K & 42.4 & 26.7 & & 87.52\% & 97.02\% \\
\rowcolor{gray!10} +SynthAlign-9K & 40.7 & 24.8 & & 87.62\% & 97.81\% \\
\rowcolor{gray!10} +SynthAlign-12K & \textbf{39.0} & \textbf{23.6} & & \textbf{88.14\%} & 97.53\% \\
\midrule
LLaVA-1.6-7B & 11.7 & 6.8 & & 86.5\% & 95.7\% \\
\rowcolor{gray!10} + SynthAlign-9K & 14.0 & 8.5 & & 84.9\% & \textbf{96.2\%} \\
\rowcolor{gray!10} + SynthAlign-12K & \textbf{10.0} & \textbf{5.9} & & \textbf{86.6\%} & 94.9\% \\
\bottomrule
\end{tabular}
}
\end{center}
\vspace{-10pt}
\caption{Comparison of hallucination evaluation results on Obj.Hal Bench and POPE on various models, and size of the synthetic dataset.}
\label{tab:objhal-bench}
\setlength{\tabcolsep}{\oldtabcolsep}
\end{table}
\\
\noindent\textbf{Highlight the Importance of Synthetic Data.} The utilization of synthetic data in our framework plays a crucial role in bridging the gap between model capabilities and real-world applications. By generating a diverse set of image-text pairs using generative models, we can simulate a wide range of scenarios that the LVLM might encounter. Despite synthetic sources, we observe broad general benchmark gains on LLaVA-1.5 and hallucination benchmark on a stronger benchmark (LLaVA-1.6). This particularly because synthetic data allows the model to learn from a richer set of examples than what might be available through limited human-annotated datasets. In other words, using synthetic data allows precise control over specific image types or scenarios, especially in domains where real data is scarce or difficult to obtain. We experiment with different dataset setups to highlight the effectiveness of synthetic data. Specifically, we sample our dataset at different sizes ranging from 2K to 12K and measure the model's hallucination rate after training on each sample. The results shown in Table \ref{tab:objhal-bench} indicate that as the size of the dataset increases, the performance improves accordingly, demonstrating the effectiveness of our framework. This also highlights that carefully constructed and selected synthetic data can benefit model training, while manage to reduce the dataset size and training compute.\\
\textbf{Issue on Synthetic Data.} In general, relying on synthesised data without careful quality control can lead to reduction in data diversity and introduce distortions. In addition, despite the benefits of using synthetic data, it often fails to capture the full complexity and nuances of real-world scenarios. The quality of generated images remains limited, sometimes producing imprecise or unrealistic visuals. Furthermore, responses generated by current SOTA open-source LVLMs can be noisy, containing errors or irrelevant information. Any biases inherent in the generative models may be propagated or even amplified in the synthetic data, negatively affecting the LVLM's fairness and reliability. Therefore, it is crucial to carefully curate and validate synthetic datasets to ensure they contribute positively to the model's training. When feasible, manual human evaluation may still be necessary to guarantee dataset quality.\\
\hspace{-1.08em}\noindent\textbf{Reward Model vs Ground-Truth Scoring Model.} We specifically analyze which model is more in line with the reward model by examining the overlap between images selected by different common methods, such as CLIP and BLIP, and examine whether they agree on the best image for a given prompt. Table \ref{common-selection} shows the percentage of common image selections among method pairs, where the Top 1 indicates the highest-ranked image agreed upon, Top 2 includes the top two shared images, and Top 3 covers the top three images.
The reward model is most in line with CLIP, showing the highest common selection percentages of 36.8\% at Top 1, 59.5\% at Top 2, and 79.13\% at Top 3 compared to other method pairs. Reward model and BLIP also show a high degree of overlap, with 31.8\% at Top 1 and 77.4\% at Top 3, indicating that reward model captures not only image-text alignment but also aspects of visual quality that BLIP may prioritize. 
\begin{table}[H]
\centering
\resizebox{0.45\textwidth}{!}{ % Adjusted to fit half the page width
\begin{tabular}{l c c c}
\toprule
\multirow{2}{*}{Method Pair} & \multicolumn{3}{c}{\textsc{Common Selection} (\%)} \\ 
\cline{2-4}
 & @ Top 1 & @ Top 2 & @ Top 3 \\
\midrule
\rowcolor{gray!10} Reward Model \& CLIP  & 36.8 & 59.5 & 79.13 \\
Reward Model \& BLIP  & 31.8 & 53.6 & 77.40 \\
CLIP \& BLIP         & 28.6 & 55.9 & 76.53 \\
All 3 Methods        & 11.8 & 34.2 & 61.87 \\
\bottomrule
\end{tabular}
    }
\caption{Percentage of common image selections among Reward Model, CLIP, and BLIP.}
\label{common-selection}
\end{table}
\begin{table}[H]
\centering
\begin{tabular}{l c c l}
\toprule
Outcome      & Count & Win Rate (\%) & \\
\midrule
\rowcolor{gray!10} RM Wins      & 53    & 58.9\%       & \rule{2.36cm}{8pt} \\
CLIP Wins    & 37    & 41.1\%       & \rule{1.64cm}{8pt} \\
Ties         & 10    & 10\%       & \rule{0.4cm}{8pt} \\
\bottomrule
\end{tabular}
\caption{LLM-as-a-Judge (GPT-4o) outcomes with counts, win rates for CLIP and Reward Model.}
\label{table:outcomes}
\end{table}
% \noindent\textbf{Effects of Prompts on Responses.} We analyzed the generated responses from the PDS-DPO framework, comparing them with outputs from the LLaVA-RLHF and RLAIF-V models. As illustrated in Figure \ref{fig:comparative_analysis}, PDS-DPO shows a distinct advantage over the other methods by exhibiting lower average similarity in both inter-response and inter-prompt distributions. This suggests that PDS-DPO produces more diverse and varied outputs, reducing redundancy and making it more adaptable to different contexts. This observation is further supported by our datasets, which consist of ten different topics, emphasizing the model's ability to handle a wide range of scenarios. The greater diversity in outputs can be beneficial in tasks requiring context-specific variations, giving our approach a potential advantage over LLaVA-RLHF and RLAIF-V.

\section{Conclusion}
In this work, we attempt vision-language post-training alignment with synthetic human-preference (image-text) data with DPO. At the core of the proposed framework is the utilization of reward models as a proxy for human preference to rate and rank the generated images and text, and hence select only the highest-rated images and text. Our experiments demonstrate favorable improvements in trustworthiness and reasoning capabilities across hallucination benchmarks and vision-language tasks, offering an efficient solution for safer LVLM deployment.\newpage
\bibliography{aaai2026}

@String(CVPR= {IEEE Conf. Comput. Vis. Pattern Recog.})

@String(ICCV= {Int. Conf. Comput. Vis.})

@String(ECCV= {Eur. Conf. Comput. Vis.})

@String(ICLR = {Int. Conf. Learn. Represent.})

@String(CVPR  = {CVPR})

@String(ICCV  = {ICCV})

@String(ECCV  = {ECCV})

@String(ICLR  = {ICLR})

@inproceedings{Alayrac2022FlamingoAV,
  title={Flamingo: a Visual Language Model for Few-Shot Learning},
  author={Jean-Baptiste Alayrac and Jeff Donahue and Pauline Luc and Antoine Miech and Iain Barr and Yana Hasson and Karel Lenc and Arthur Mensch and Katie Millican and Malcolm Reynolds and Roman Ring and Eliza Rutherford and Serkan Cabi and Tengda Han and Zhitao Gong and Sina Samangooei and Marianne Monteiro and Jacob Menick and Sebastian Borgeaud and Andy Brock and Aida Nematzadeh and Sahand Sharifzadeh and Mikolaj Binkowski and Ricardo Barreira and Oriol Vinyals and Andrew Zisserman and Karen Simonyan},
  booktitle={NeurIPS},
  year={2022}
}

@inproceedings{BLIP2,
title={BLIP-2: Bootstrapping Language-Image Pre-training with Frozen Image Encoders and Large Language Models}, 
author={Junnan Li and Dongxu Li and Silvio Savarese and Steven Hoi},
year={2023},
booktitle={ICML} 
}

@inproceedings{MiniGPT4,
title={MiniGPT-4: Enhancing Vision-Language Understanding with Advanced Large Language Models}, 
author={Deyao Zhu and Jun Chen and Xiaoqian Shen and Xiang Li and Mohamed Elhoseiny},
year={2023},
booktitle={ICLR}
}

@inproceedings{liu2023visualinstructiontuning,
title={Visual Instruction Tuning}, 
author={Haotian Liu and Chunyuan Li and Qingyang Wu and Yong Jae Lee},
year={2023},
booktitle={NeurIPS}
}

@inproceedings{Deita,
title={What Makes Good Data for Alignment? A Comprehensive Study of Automatic Data Selection in Instruction Tuning}, 
author={Wei Liu and Weihao Zeng and Keqing He and Yong Jiang and Junxian He},
year={2024},
booktitle={ICLR}
}

@inproceedings{Alpagasus,
title={AlpaGasus: Training A Better Alpaca with Fewer Data}, 
author={Lichang Chen and Shiyang Li and Jun Yan and Hai Wang and Kalpa Gunaratna and Vikas Yadav and Zheng Tang and Vijay Srinivasan and Tianyi Zhou and Heng Huang and Hongxia Jin},
year={2024},
booktitle={ICLR}
}

@inproceedings{LIMA,
title={LIMA: Less Is More for Alignment}, 
author={Chunting Zhou and Pengfei Liu and Puxin Xu and Srini Iyer and Jiao Sun and Yuning Mao and Xuezhe Ma and Avia Efrat and Ping Yu and Lili Yu and Susan Zhang and Gargi Ghosh and Mike Lewis and Luke Zettlemoyer and Omer Levy},
year={2023},
booktitle={NeurIPS}
}

@article{Bunny,
title={Efficient Multimodal Learning from Data-centric Perspective}, 
author={Muyang He and Yexin Liu and Boya Wu and Jianhao Yuan and Yueze Wang and Tiejun Huang and Bo Zhao},
year={2024},
journal={arXiv Preprint arXiv: 2402.11530},
}

@inproceedings{TLDR,
title={Too Large; Data Reduction for Vision-Language Pre-Training}, 
author={Alex Jinpeng Wang and Kevin Qinghong Lin and David Junhao Zhang and Stan Weixian Lei and Mike Zheng Shou},
year={2023},
booktitle={ICCV}
}

@inproceedings{chen2023sharegpt4v,
title={ShareGPT4V: Improving Large Multi-Modal Models with Better Captions},
author={Chen, Lin and Li, Jisong and Dong, Xiaoyi and Zhang, Pan and He, Conghui and Wang, Jiaqi and Zhao, Feng and Lin, Dahua},
booktitle={ECCV},
year={2023}
}

@article{chen2024allava,
title={ALLaVA: Harnessing GPT4V-synthesized Data for A Lite Vision-Language Model}, 
author={Guiming Hardy Chen and Shunian Chen and Ruifei Zhang and Junying Chen and Xiangbo Wu and Zhiyi Zhang and Zhihong Chen and Jianquan Li and Xiang Wan and Benyou Wang},
year={2024},
journal={arXiV Preprint arXiv: 2402.11684}
}

@article{li2024recaption,
title={What If We Recaption Billions of Web Images with LLaMA-3?}, 
author={Xianhang Li and Haoqin Tu and Mude Hui and Zeyu Wang and Bingchen Zhao and Junfei Xiao and Sucheng Ren and Jieru Mei and Qing Liu and Huangjie Zheng and Yuyin Zhou and Cihang Xie},
journal={arXiv preprint arXiv:2406.08478},
year={2024}
}

@article{zhao2023svit,
title={SVIT: Scaling up Visual Instruction Tuning}, 
author={Zhao, Bo and Wu, Boya and He, Muyang and Huang, Tiejun},
journal={arXiv preprint arXiv:2307.04087},
year={2023}
}

@inproceedings{yu2023capsfusion,
title={CapsFusion: Rethinking Image-Text Data at Scale},
author={Yu, Qiying and Sun, Quan and Zhang, Xiaosong and Cui, Yufeng and Zhang, Fan and Cao, Yue and Wang, Xinlong and Liu, Jingjing},
booktitle={CVPR},
year={2024}
}

@inproceedings{lai2024veclip,
title={VeCLIP: Improving CLIP Training via Visual-enriched Captions}, 
author={Zhengfeng Lai and Haotian Zhang and Bowen Zhang and Wentao Wu and Haoping Bai and Aleksei Timofeev and Xianzhi Du and Zhe Gan and Jiulong Shan and Chen-Nee Chuah and Yinfei Yang and Meng Cao},
year={2024},
booktitle={ECCV}
}

@article{sun2023llavarlhf,
  title={Aligning large multimodal models with factually augmented rlhf},
  author={Sun, Zhiqing and Shen, Sheng and Cao, Shengcao and Liu, Haotian and Li, Chunyuan and Shen, Yikang and Gan, Chuang and Gui, Liang-Yan and Wang, Yu-Xiong and Yang, Yiming and others},
  journal={arXiv preprint arXiv:2309.14525},
  year={2023}
}

@article{2023silkie,
  author={Lei Li and Zhihui Xie and Mukai Li and Shunian Chen and Peiyi Wang and Liang Chen and  Yazheng Yang and  Benyou Wang and  Lingpeng Kong},
  title={Silkie: Preference Distillation for Large Visual Language Models},
  journal={arXiv preprint arXiv:2312.10665},
  year={2023}
}

@article{yu2024rlaif,
  title={Rlaif-v: Aligning mllms through open-source ai feedback for super gpt-4v trustworthiness},
  author={Yu, Tianyu and Zhang, Haoye and Yao, Yuan and Dang, Yunkai and Chen, Da and Lu, Xiaoman and Cui, Ganqu and He, Taiwen and Liu, Zhiyuan and Chua, Tat-Seng and others},
  journal={arXiv preprint arXiv:2405.17220},
  year={2024}
}

@inproceedings{yu2024rlhf,
  title={Rlhf-v: Towards trustworthy mllms via behavior alignment from fine-grained correctional human feedback},
  author={Yu, Tianyu and Yao, Yuan and Zhang, Haoye and He, Taiwen and Han, Yifeng and Cui, Ganqu and Hu, Jinyi and Liu, Zhiyuan and Zheng, Hai-Tao and Sun, Maosong and others},
  booktitle={CVPR},
  pages={13807--13816},
  year={2024}
}

@article{zhou2024aligning,
  title={Aligning modalities in vision large language models via preference fine-tuning},
  author={Zhou, Yiyang and Cui, Chenhang and Rafailov, Rafael and Finn, Chelsea and Yao, Huaxiu},
  journal={arXiv preprint arXiv:2402.11411},
  year={2024}
}

@inproceedings{instructblip,
title={InstructBLIP: Towards General-purpose Vision-Language Models with Instruction Tuning}, 
author={Wenliang Dai and Junnan Li and Dongxu Li and Anthony Meng Huat Tiong and Junqi Zhao and Weisheng Wang and Boyang Li and Pascale Fung and Steven Hoi},
year={2023},
booktitle={NeurIPS} 
}

@article{mckinzie2024mm1,
  title={Mm1: Methods, analysis \& insights from multimodal llm pre-training},
  author={McKinzie, Brandon and Gan, Zhe and Fauconnier, Jean-Philippe and Dodge, Sam and Zhang, Bowen and Dufter, Philipp and Shah, Dhruti and Du, Xianzhi and Peng, Futang and Weers, Floris and others},
  journal={arXiv pPreprint arXiv:2403.09611},
  year={2024}
}

@article{laurenccon2024matters,
  title={What matters when building vision-language models?},
  author={Lauren{\c{c}}on, Hugo and Tronchon, L{\'e}o and Cord, Matthieu and Sanh, Victor},
  journal={arXiv preprint arXiv:2405.02246},
  year={2024}
}

@inproceedings{rafailov2024direct,
  title={Direct preference optimization: Your language model is secretly a reward model},
  author={Rafailov, Rafael and Sharma, Archit and Mitchell, Eric and Manning, Christopher D and Ermon, Stefano and Finn, Chelsea},
  journal={NeurIPS},
  volume={36},
  year={2024}
}

@article{ouyang2022training,
  title={Training language models to follow instructions with human feedback},
  author={Ouyang, Long and Wu, Jeffrey and Jiang, Xu and Almeida, Diogo and Wainwright, Carroll and Mishkin, Pamela and Zhang, Chong and Agarwal, Sandhini and Slama, Katarina and Ray, Alex and others},
  journal={NeurIPS},
  volume={35},
  pages={27730--27744},
  year={2022}
}

@article{schulman2017proximal,
  title={Proximal policy optimization algorithms},
  author={Schulman, John and Wolski, Filip and Dhariwal, Prafulla and Radford, Alec and Klimov, Oleg},
  journal={arXiv preprint arXiv:1707.06347},
  year={2017}
}

@article{zhang2024lmmsevalrealitycheckevaluation,
      title={LMMs-Eval: Reality Check on the Evaluation of Large Multimodal Models}, 
      author={Kaichen Zhang and Bo Li and Peiyuan Zhang and Fanyi Pu and Joshua Adrian Cahyono and Kairui Hu and Shuai Liu and Yuanhan Zhang and Jingkang Yang and Chunyuan Li and Ziwei Liu},
      year={2024},
      journal={arXiv preprint arXiv:2407.12772}
}

@article{xu2024imagereward,
  title={Imagereward: Learning and evaluating human preferences for text-to-image generation},
  author={Xu, Jiazheng and Liu, Xiao and Wu, Yuchen and Tong, Yuxuan and Li, Qinkai and Ding, Ming and Tang, Jie and Dong, Yuxiao},
  journal={NeurIPS},
  volume={36},
  year={2024}
}

@article{wang2024interpretable,
  title={Interpretable Preferences via Multi-Objective Reward Modeling and Mixture-of-Experts},
  author={Wang, Haoxiang and Xiong, Wei and Xie, Tengyang and Zhao, Han and Zhang, Tong},
  journal={arXiv preprint arXiv:2406.12845},
  year={2024}
}

@article{bai2022constitutional,
  title={Constitutional ai: Harmlessness from ai feedback},
  author={Bai, Yuntao and Kadavath, Saurav and Kundu, Sandipan and Askell, Amanda and Kernion, Jackson and Jones, Andy and Chen, Anna and Goldie, Anna and Mirhoseini, Azalia and McKinnon, Cameron and others},
  journal={arXiv preprint arXiv:2212.08073},
  year={2022}
}

@inproceedings{liu2024improved,
  title={Improved baselines with visual instruction tuning},
  author={Liu, Haotian and Li, Chunyuan and Li, Yuheng and Lee, Yong Jae},
  booktitle={CVPR},
  pages={26296--26306},
  year={2024}
}

@article{Qwen-VL,
  title={Qwen-VL: A Versatile Vision-Language Model for Understanding, Localization, Text Reading, and Beyond},
  author={Bai, Jinze and Bai, Shuai and Yang, Shusheng and Wang, Shijie and Tan, Sinan and Wang, Peng and Lin, Junyang and Zhou, Chang and Zhou, Jingren},
  journal={arXiv preprint arXiv:2308.12966},
  year={2023}
}

@article{li2023evaluating,
  title={Evaluating object hallucination in large vision-language models},
  author={Li, Yifan and Du, Yifan and Zhou, Kun and Wang, Jinpeng and Zhao, Wayne Xin and Wen, Ji-Rong},
  journal={arXiv preprint arXiv:2305.10355},
  year={2023}
}

@misc{zhao2023hallucinations,
      title={Beyond Hallucinations: Enhancing LVLMs through Hallucination-Aware Direct Preference Optimization}, 
      author={Zhiyuan Zhao and Bin Wang and Linke Ouyang and Xiaoyi Dong and Jiaqi Wang and Conghui He},
      year={2023},
      eprint={2311.16839},
      archivePrefix={arXiv},
      primaryClass={cs.CV}
}

@inproceedings{ouali2024clip,
  title={CLIP-DPO: Vision-Language Models as a Source of Preference for Fixing Hallucinations in LVLMs},
  author={Ouali, Yassine and Bulat, Adrian and Martinez, Brais and Tzimiropoulos, Georgios},
  booktile={ECCV},
  year={2024}
}

@article{bradley1952rank,
  title={Rank analysis of incomplete block designs: I. The method of paired comparisons},
  author={Bradley, Ralph Allan and Terry, Milton E},
  journal={Biometrika},
  volume={39},
  number={3/4},
  pages={324--345},
  year={1952},
  publisher={JSTOR}
}

@article{rohrbach2018object,
  title={Object hallucination in image captioning},
  author={Rohrbach, Anna and Hendricks, Lisa Anne and Burns, Kaylee and Darrell, Trevor and Saenko, Kate},
  journal={arXiv preprint arXiv:1809.02156},
  year={2018}
}

@article{wang2023llm,
  title={An LLM-free Multi-dimensional Benchmark for MLLMs Hallucination Evaluation},
  author={Wang, Junyang and Wang, Yuhang and Xu, Guohai and Zhang, Jing and Gu, Yukai and Jia, Haitao and Yan, Ming and Zhang, Ji and Sang, Jitao},
  journal={arXiv preprint arXiv:2311.07397},
  year={2023}
}

@InProceedings{Hataya_2023_ICCV,
    author    = {Hataya, Ryuichiro and Bao, Han and Arai, Hiromi},
    title     = {Will Large-scale Generative Models Corrupt Future Datasets?},
    booktitle = {ICCV},
    year      = {2023},
    pages     = {20555-20565}
}

@article{fu2023mme,
  title={MME: A Comprehensive Evaluation Benchmark for Multimodal Large Language Models},
  author={Fu, Chaoyou and Chen, Peixian and Shen, Yunhang and Qin, Yulei and Zhang, Mengdan and Lin, Xu and Yang, Jinrui and Zheng, Xiawu and Li, Ke and Sun, Xing and others},
  journal={arXiv preprint arXiv:2306.13394},
  year={2023}
}

@article{shumailov2023curse,
  title={The curse of recursion: Training on generated data makes models forget},
  author={Shumailov, Ilia and Shumaylov, Zakhar and Zhao, Yiren and Gal, Yarin and Papernot, Nicolas and Anderson, Ross},
  journal={arXiv preprint arXiv:2305.17493},
  year={2023}
}

@article{lu2022learn,
  title={Learn to explain: Multimodal reasoning via thought chains for science question answering},
  author={Lu, Pan and Mishra, Swaroop and Xia, Tanglin and Qiu, Liang and Chang, Kai-Wei and Zhu, Song-Chun and Tafjord, Oyvind and Clark, Peter and Kalyan, Ashwin},
  journal={NeurIPS},
  volume={35},
  pages={2507--2521},
  year={2022}
}

@article{chen2023shikra,
  title={Shikra: Unleashing Multimodal LLM's Referential Dialogue Magic},
  author={Chen, Keqin and Zhang, Zhao and Zeng, Weili and Zhang, Richong and Zhu, Feng and Zhao, Rui},
  journal={arXiv preprint arXiv:2306.15195},
  year={2023}
}

@article{radford2021learningtransferablevisualmodels,
      title={Learning Transferable Visual Models From Natural Language Supervision}, 
      author={Alec Radford and Jong Wook Kim and Chris Hallacy and Aditya Ramesh and Gabriel Goh and Sandhini Agarwal and Girish Sastry and Amanda Askell and Pamela Mishkin and Jack Clark and Gretchen Krueger and Ilya Sutskever},
      year={2021},
      journal={arXiv preprint arXiv:2103.00020},
}

@article{hu2021lora,
  title={Lora: Low-rank adaptation of large language models},
  author={Hu, Edward J and Shen, Yelong and Wallis, Phillip and Allen-Zhu, Zeyuan and Li, Yuanzhi and Wang, Shean and Wang, Lu and Chen, Weizhu},
  journal={arXiv preprint arXiv:2106.09685},
  year={2021}
}

@article{yu2023mm,
  title={Mm-vet: Evaluating large multimodal models for integrated capabilities},
  author={Yu, Weihao and Yang, Zhengyuan and Li, Linjie and Wang, Jianfeng and Lin, Kevin and Liu, Zicheng and Wang, Xinchao and Wang, Lijuan},
  journal={arXiv preprint arXiv:2308.02490},
  year={2023}
}

@inproceedings{chen2024internvl,
  title={Internvl: Scaling up vision foundation models and aligning for generic visual-linguistic tasks},
  author={Chen, Zhe and Wu, Jiannan and Wang, Wenhai and Su, Weijie and Chen, Guo and Xing, Sen and Zhong, Muyan and Zhang, Qinglong and Zhu, Xizhou and Lu, Lewei and others},
  booktitle={Proceedings of the IEEE/CVF conference on computer vision and pattern recognition},
  pages={24185--24198},
  year={2024}
}

@article{gunasekar2023textbooks,
  title={Textbooks are all you need},
  author={Gunasekar, Suriya and Zhang, Yi and Aneja, Jyoti and Mendes, Caio C{\'e}sar Teodoro and Del Giorno, Allie and Gopi, Sivakanth and Javaheripi, Mojan and Kauffmann, Piero and de Rosa, Gustavo and Saarikivi, Olli and others},
  journal={arXiv preprint arXiv:2306.11644},
  year={2023}
}

@inproceedings{gurari2018vizwiz,
  title={Vizwiz grand challenge: Answering visual questions from blind people},
  author={Gurari, Danna and Li, Qing and Stangl, Abigale J and Guo, Anhong and Lin, Chi and Grauman, Kristen and Luo, Jiebo and Bigham, Jeffrey P},
  booktitle={Proceedings of the IEEE conference on computer vision and pattern recognition},
  pages={3608--3617},
  year={2018}
}

@article{li2023seed,
  title={Seed-bench: Benchmarking multimodal llms with generative comprehension},
  author={Li, Bohao and Wang, Rui and Wang, Guangzhi and Ge, Yuying and Ge, Yixiao and Shan, Ying},
  journal={arXiv preprint arXiv:2307.16125},
  year={2023}
}

@inproceedings{yue2024mmmu,
  title={Mmmu: A massive multi-discipline multimodal understanding and reasoning benchmark for expert agi},
  author={Yue, Xiang and Ni, Yuansheng and Zhang, Kai and Zheng, Tianyu and Liu, Ruoqi and Zhang, Ge and Stevens, Samuel and Jiang, Dongfu and Ren, Weiming and Sun, Yuxuan and others},
  booktitle={Proceedings of the IEEE/CVF Conference on Computer Vision and Pattern Recognition},
  pages={9556--9567},
  year={2024}
}

\end{document}